\pdfoutput=1
\documentclass[11pt]{article}
\usepackage{acl}

\usepackage{times}
\usepackage{latexsym}
\usepackage[T1]{fontenc}
\usepackage[ruled,vlined]{algorithm2e}

\usepackage{tabularx}
\usepackage{booktabs}
\usepackage[utf8]{inputenc}
\usepackage{microtype}
\usepackage{inconsolata}
\usepackage{graphicx}
\usepackage{multirow}
\usepackage{float}

\newcommand{\emoclass}{\texttt{$\langle$em$\rangle$}\xspace}

\title{MOPO: Multi-Objective Prompt Optimization for\linebreak Affective Text Generation}

\author{Yarik Menchaca Resendiz$^{1,2}$ \and Roman Klinger$^{2}$ \\
  $^{1}$Institut f\"ur Maschinelle Sprachverarbeitung, University of Stuttgart, Germany\\
  $^{2}$Fundamentals of Natural Language Processing, University of Bamberg, Germany\\
  \texttt{\{yarik.menchaca-resendiz,roman.klinger\}@uni-bamberg.de}
  }

\begin{document}
\maketitle
\begin{abstract}
  How emotions are expressed depends on the context and domain. On
  X (formerly Twitter), for instance, an author might simply use the hashtag
  \texttt{\#anger}, while in a news headline, emotions are typically
  written in a more polite, indirect manner. To enable conditional
  text generation models to create emotionally connotated texts that
  fit a domain, users need to have access to a parameter that
  allows them to choose the appropriate way to express an emotion. To
  achieve this, we introduce MOPO, a Multi-Objective Prompt
  Optimization methodology. MOPO optimizes prompts according to
  multiple objectives (which correspond here to the output
  probabilities assigned by emotion classifiers trained for different
  domains). In contrast to single objective optimization, MOPO outputs
  a set of prompts, each with a different weighting of the multiple
  objectives. Users can then choose the most appropriate prompt for
  their context.  We evaluate MOPO using three objectives, determined
  by various domain-specific emotion classifiers. MOPO improves
  performance by up to 15 pp across all objectives with a minimal loss
  (1--2 pp) for any single objective compared to single-objective
  optimization. These minor performance losses are offset by a broader
  generalization across multiple objectives -- which is not possible
  with single-objective optimization. Additionally, MOPO reduces
  computational requirements by simultaneously optimizing for multiple
  objectives, eliminating separate optimization procedures for each
  objective.
\end{abstract}

\section{Introduction}

\begin{figure}[t]
  \includegraphics[width=\columnwidth]{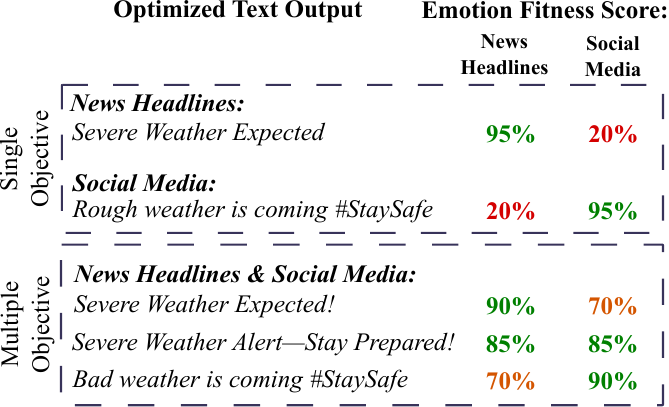}
  \caption{Examples of prompt-based generated text. The prompts are
    optimized for two conflicting objectives: News Headlines and
    Social Media. The Emotion Fitness Score evaluates how well the text
    fulfills each objective. In the Single Objective section, prompts
    are optimized either for News Headlines (high score for news) or
    Social Media (high score for social media), leading to lower
    fitness scores in the other category. In contrast, Multi-Objective
    prompts optimize for both News Headlines and Social Media
    simultaneously, generating a range of high-performing
    options. Users can select the best-performing prompt for each
    objective or choose a balanced option (e.g., \textit{``Severe
      Weather Alert -- Stay Prepared''}, which fits 85\% across all
    objectives).}
  \label{fig:basic_example}
\end{figure}

Large language models (LLMs) have improved system performances on many
natural language processing (NLP) tasks. The standard approach to find
prompts is either manual prompt engineering or automatic prompt
optimization with some annotated data. In the case of prompt
optimization, it is however difficult to consider all relevant
aspects: Real-world applications often demand prompts that satisfy
multiple requirements (objectives) simultaneously. For instance, in healthcare systems, prompts must balance clarity and accuracy (factuality) to provide information that is both understandable and reliable. However, simplifying medical information for clarity might compromise medical accuracy. Similarly, in affective text generation (our use case), a newspaper headline is usually formal,
while the same meaning would be communicated in a more informal way in
social media. Figure \ref{fig:basic_example} shows an example,
including an output that would be acceptable across domains. Automatic
prompt optimization can lead to a well-performing prompt for the
domain it has been optimized for, but it might not generalize well to
other domains.

\begin{figure}
  \centering
  \includegraphics{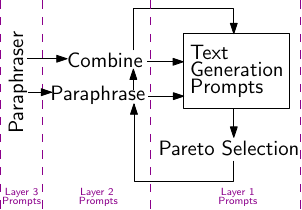}
  \caption{Three layers of prompts in our MOPO approach for multi-objective prompt optimization for affective text generation.}
  \label{fig:optimizationlayers}
\end{figure}

To enable end-users to select their desired weighting across multiple
domains without retraining the prompts, we introduce the
Multi-Objective Prompt Optimization (MOPO) method. It consists of a
three-layer optimization model, two of which are self-optimizing (see
Figure~\ref{fig:optimizationlayers}).
Each layer corresponds to a set of prompts and specific tasks in the
optimization process. Layer-1 consists of prompts that solve the task
at hand: affective text generation (e.g., \textit{``Write a text that
  expresses Joy''}). Layer-2 consists of prompts that change the set
of Layer-1 prompts, by paraphrasing and combing them into new prompts
(e.g., \textit{``Paraphrase\ldots''} or \textit{``Mix the two prompts
  \ldots\ into a new single prompt.''} ). Layer-3 changes Layer-2
prompts such that they are potentially more effective in optimizing
Layer-1 prompts. Layer-3 is not iteratively optimized.  MOPO uses
Pareto optimization to explore trade-offs between multiple objectives
within the Layer-1 prompts by applying the Non-dominated Sorting
Genetic Algorithm II \cite[NSGA-II,][]{deb2000fast}\footnote{The code
  and resources can be found at
  \url{https://www.uni-bamberg.de/en/nlproc/resources/mopo/}}.
 
To understand the properties of MOPO, we answer the following research
questions: \textit{``RQ1: How does single-objective prompt
  optimization for affective text generation compare to
  multi-objective prompt optimization?''}, \textit{``RQ2: How do
  paraphrasing and combining prompts affect the performance of the overall
  optimization procedure?''}, and \textit{``RQ3: How does
  multi-objective optimization impact the quality of the generated
  texts?''}. 

\section{Related Work}
\label{sec:relatedwork}
\subsection{Affective Text Generation}
Research on conditional language generation has predominantly focused
on sentiment polarity \citep{zhang2019emotional, maqsud2015synthetic,
  niu2018polite} and generating text based on topics
\cite{orbach2020facts2story,chan2020cocon}. Among the few studies
addressing emotion conditions, Affect-LM \cite{ghosh-etal-2017-affect}
stands out as a language model for crafting conversational texts. In
the area of dialogue systems, the Emotional Chatting Machine
\citep{zhou2018emotional} integrates modules for abstract emotion
representation, emotion state transitions, and uses an external
emotion lexicon. EmoDS \citep{song2019generating} generates responses
conveying specific emotions through either direct input or context,
using a sequence-level emotion
classifier. \citet{colombo-etal-2019-affect} introduces a GPT-2-based
\citep{radford2019language} framework that combines classifiers with
emotion and topic lexicons for conditioned outputs. The Multi-turn
Emotional Conversation Model \citep[MECM]{cui2022modeling} enhances
conversation by maintaining emotional continuity. Furthermore,
\citet{menchaca-resendiz-klinger-2023-affective} demonstrate that
incorporating appraisal alongside emotion conditions enables
fine-grained control of emotion generated text. Further, they show how
prompts can be automatically optimized for affective text generation
\citep{resendiz-klinger-2023-emotion}.

\subsection{Multi-Objective Optimization}

Genetic algorithms (GAs), introduced by
\citet{holland_adaptation_1975}, are used in various optimization
tasks due to their ability to explore large and complex solution
spaces \citep{goldberg89, Mitchell1996Introduction}.  This exploration
is achieved by introducing randomized changes to individual solutions
(mutation) and combining traits from two-parent solutions (crossover)
to create new candidates. Genetic evolutionary optimization handles
multiple solutions at the same time, and is therefore a
straight-forward candidate for extension to multi-objective
optimization. Here, each solution has a different weighting of
multiple objectives, offering a selection to end users. Prominent
instances of such multi-objective optimization methods are NSGA
\citep{srinivas1994multiobjective}, NSGA-II \citep{NSGA_2}, and
NSGA-III \citep{nsga-iii}, which use Pareto optimization
\citep{ParetoManual} to rank solutions based on competing objectives.

In NLP, GAs have been applied to tasks such as machine translation,
where \citet{jon-bojar-2023-breeding} explored modifications to
mutation and crossover processes.  In a similar context,
\citet{huang-etal-2023-towards} used Pareto optimization to manage
trade-offs between two languages.
\citet{liu-etal-2022-towards-efficient} introduced an evaluation
framework that utilizes the Pareto Frontier to assess performance
across various language understanding tasks. This work showcases the
utility of Pareto optimization in enhancing language models'
efficiency and efficacy.

\subsection{Prompt Optimization}

LLMs have demonstrated the ability to solve tasks in zero- or few-shot
learning settings via prompting \citep{semnani-etal-2023-wikichat,
  lin-etal-2022-shot}. These prompts include instructions to guide the
model's text generation. For example, in text classification
\cite{hu-etal-2022-knowledgeable, gu-etal-2022-ppt}, they combine the
instruction with a class label (e.g., \textit{``Tag the following text
  as positive or negative \ldots''}). Summarization prompts mention
keywords like \textit{``TL;DR''} or \textit{``summarize''}
\cite{radford2019language, narayan-etal-2021-planning}. Machine
translation prompts \cite{raffel2019exploring} specify the languages
to translate between (e.g., \textit{``Translate English to German''}).

While manual prompt development can be successful, automatic prompt
optimization is crucial for overcoming limited adaptability and user
subjectivity. AutoPrompt \cite{autoprompt:emnlp20} suggests a
``fill-in-the-blanks'' method with gradient-guided
search. \citet{resendiz-klinger-2023-emotion} introduce an iterative
method for automatic prompt optimization in emotion-conditioned text
generation, which modifies prompts by adding, replacing, or removing
tokens. OpenPrompt \cite{ding2021openprompt} provides tools for prompt
training through templates and
verbalizers. \citet{deng-etal-2022-rlprompt} employ reinforcement
learning to discover effective prompt variation tactics. Promptbreeder
\citep{fernando2024promptbreeder} uses self-referential optimization
of a group of task prompts. We build on top of their work and extend
it to text generation. Further, we include Pareto optimization, which
has, so far, not been used in any prompt optimization task.

\section{MOPO}
In the following section, we introduce our Multi-Objective Prompt Optimization
method MOPO for emotion-conditioned text generation. It uses prompts
for text generation, which are optimized with Pareto optimization
following multiple objectives. We refer to these \textit{task-specific} prompts
as Layer-1 prompts. The variations in the set of prompts are induced by
paraphrasing them (in GA terminology: mutation) and combining them (in
GA terminology: crossover). These variations are performed via
prompting as well (we refer to these prompts as Layer-2 prompts). The
Layer-2 prompts that perform the variations on Layer-1 prompts are
further optimized by fixed prompts which we refer as Layer-3. The
selection of Layer-1 prompts follows multiple objective functions --
Layer-2 prompts are selected based on their contribution to the
success of Layer-1 prompts. This intuitive understanding, visualized
in Figure~\ref{fig:optimizationlayers}, is explained more formally in
the next section.

\begin{algorithm}[t]
  \footnotesize
  \SetKwInOut{Input}{Input} \SetKwInOut{Output}{Output}
  \Input{Seed Prompts $\mathbf{SP}$, \linebreak Combine Prompts $\mathbf{P}_\textit{c}$, \linebreak Paraphrase Prompts $\mathbf{P}_\textit{p}$, \linebreak   Generations $I$, \linebreak Generation size $G$, \linebreak Max Chromosomes per Breeding $C$}
  \Output{Optimized Prompts $\mathbf{P}_\textit{opt}$}
  \SetKwFunction{Add}{Add}
  
  $\mathbf{P}_\textit{opt} \gets \mathbf{SP}$\;
  $i \gets 0$\;
  $\mathbf{P}_\textit{cands} \gets \{\}$\;
  
  \While{$i < I$}{
    $\mathbf{P}_\textit{pop} \gets \mathbf{P}_\textit{opt}$\; 
    
    $\mathbf{P}_\textit{pop}, \mathbf{P}_\textit{c} \,+\kern-3pt=\textit{Combine}(\mathbf{P}_\textit{pop},\mathbf{P}_\textit{c}, C)$\;
    	
    $\mathbf{P}_\textit{pop}, \mathbf{P}_\textit{p} \,+\kern-3pt=\textit{Paraphrase}(\mathbf{P}_\textit{pop}, \mathbf{P}_\textit{p}, C)$\;   
    
    $\mathbf{T}_\textit{pop} \gets \textit{TextGeneration}(\mathbf{P}_\textit{pop}) $\;
    
    $\mathbf{T}_\textit{eval} \gets \textit{FitnessEvaluation}(\mathbf{T}_\textit{pop})$\;
    
    $\mathbf{P}_\textit{opt} \gets \textit{ParetoSelection}(\mathbf{T}_\textit{eval}, G)$ \;
    
    $\mathbf{P}_\textit{cands} +\kern-3pt= \mathbf{P}_\textit{opt}$ \;
    
    $ \mathbf{P}_\textit{c} \gets \textit{CombinePromptSelection}(\mathbf{P}_\textit{c})$ \;
    $ \mathbf{P}_\textit{p} \gets \textit{ParaphrasePromptSelection}(\mathbf{P}_\textit{p})$ \;
    
    $i \gets i + 1$\;
  }
  $\mathbf{P}_\textit{opt} \gets \textrm{ParetoSelection}(\mathbf{P}_\textit{cands}, G)$\;
  
  \Return $P_\textit{opt}$\;
  \caption{MOPO}
  \label{algorithm:pseudocode}
\end{algorithm}

\subsection{Algorithm}
\label{sub:algorithm}
The iterative process (Algorithm \ref{algorithm:pseudocode}) optimizes
a set of \textit{task-specific} prompts (Layer-1, e.g., ``Write a text
that expresses Joy''). Initially, the optimized prompts
$\mathbf{P}\textit{opt}$ are the seed \textit{task-specific} prompts
$\mathbf{SP}$. Each generation starts by treating the current prompts
to be optimized ($\mathbf{P}\textit{opt}$) as the full prompt
population ($\mathbf{P}\textit{pop}$). We then expand
$\mathbf{P}\textit{pop}$ by applying the operations \textit{Combine}
and \textit{Paraphrase} (Section \ref{sub:genetic_operations}).
Next, we use a pre-trained language model to generate $n$ texts for each \textit{task-specific} prompt ($T\textit{pop}$, e.g., ``I like to eat tacos'', Section \ref{sub: text_generation}). The performance of each $\mathbf{P}\textit{pop}$ is evaluated by the \textit{FitnessEvaluation} function (Section \ref{sub:fitness_evaluation}), based on the texts it generates ($T\textit{pop}$).

The top $G$ \textit{task-specific} prompts are selected from the current
generation using non-dominated sorting within the
\textit{ParetoSelection}, forming the next generation
$\mathbf{P}_\textit{opt}$.  Finally, we optimize Layer-2 prompts
($\mathbf{P}\textit{c}$ and $\mathbf{P}\textit{p}$) to make them more
effective and adaptable in the operations \textit{Combine} and
\textit{Paraphrase}. This is done by selecting the Layer-2 prompts
that contributed to the best Layer-1 results across all objectives,
using \textit{CombinePromptSelection} and
\textit{ParaphrasePromptSelection}.

\subsection{Genetic Operations}
\label{sub:genetic_operations}
\paragraph{Combine.} 
We pair the best \textit{task-specific} prompts from each
objective to create prompts that better fulfill both objectives
simultaneously.\footnote{For example, combining $P_\textit{m}$
  (\textit{``Write a polite text expressing Joy''}) and $P_\textit{n}$
  (\textit{``Write a text expressing Joy in less than 20 words''}) can
  result in the prompt \textit{“Write a short and polite text
    expressing joy”}.} Algorithm \ref{algorithm: Crossover} first
paraphrases $\mathbf{P}_\textit{c}$ (the Layer-2 prompts that are used
to combine multiple Layer-1 prompts) using Layer-3 prompts (fixed
prompts, Table \ref{tab:fix_prompts}), with the aim of optimizing not
only the \textit{task-specific} prompts $\mathbf{P}_\textit{pop}$ but
also the $\mathbf{P}_\textit{c}$ in each
iteration. \textit{PairSample} selects all pair combinations
$P_\textit{m}, P_\textit{n}$ of the best prompts from each
objective\footnote{Selection is based on the
  \textit{FitnessEvaluation} from the previous generation or is random
  if $i = 0$.}. Finally, we generate $C$ new prompts for each prompt
in $\mathbf{P}_\textit{c}$ for each pair $P_\textit{m}, P_\textit{n}$.

\begin{algorithm}[t]
  \footnotesize
  \SetKwInOut{Input}{Input} \SetKwInOut{Output}{Output}
  \Input{Parent Prompts $\mathbf{P}_\textit{pop}$,
  	\linebreak Combine Prompts $\mathbf{P}_\textit{c}$,
  	\linebreak Max Chromosomes per Breeding $C$}
  \Output{Combined Prompts $\mathbf{P}_\textit{c-pop}$,
  	\linebreak Paraphrased Combine-Prompts $\mathbf{P}_\textit{c}$}
  	\SetKwFunction{Add}{Add}
  	
  	$\mathbf{P}_\textit{c} \gets \textit{Paraphrase}(\mathbf{P}_\textit{c}) $\;
  	\For{$P_\textit{m}, P_\textit{n}$ $\mathbf{in}$ $\textit{PairSample}(\mathbf{P}_\textit{pop})$}
  	{
  	$\mathbf{P}_\textit{c-pop} +\kern-3pt=  \textrm{PromptCombine}(P_\textit{m}, P_\textit{n}, \mathbf{P}_\textit{c}, C)$\;
  	}   
  \SetKwFunction{Add}{Add}
  
  \Return $\mathbf{P}_\textit{c-pop}$, $\mathbf{P}_\textit{c}$\;
  \caption{Combine}
  \label{algorithm: Crossover}
\end{algorithm}

\paragraph{Paraphrase.} We paraphrase each \textit{task-specific} prompt
$\mathbf{P}_\textit{m}$ within $\mathbf{P}_\textit{pop}$
individually. Analogous to the \textit{Combine} operation, we
paraphrase $\mathbf{P}_\textit{p}$ with the same fixed set of
prompts. As shown in Algorithm \ref{algorithm:paraphrase}, we perform
two separate paraphrasing steps: (1) Sentence level
(\textit{SentenceParaphrase}), which uses each paraphrase prompt in
$\mathbf{P}_\textit{c}$ (e.g., \textit{``Paraphrase the following
  sentence: \ldots''}) to generate $C$ new prompts for
$\mathbf{P}_\textit{m}$. (2) Word level (\textit{WordParaphrase}),
which involves three operations, one at a time: \textit{Addition} adds
the most probable token at any position within the prompt, including
both the beginning and the end, based on a masked pre-trained model
(e.g., RoBERTa). \textit{Removal} deletes a token from the prompt.
\textit{Replacement} exchanges a token by the most probable
token\footnote{\textit{Addition} and \textit{Replacement} use the
  \textit{$\langle$mask$\rangle$} token.}.

\begin{algorithm}[t]
  \footnotesize
  \SetKwInOut{Input}{Input} \SetKwInOut{Output}{Output}
  \Input{Parent Prompts $\mathbf{P}_\textit{pop}$,
    \linebreak Paraphrase Prompts $\mathbf{P}_\textit{p}$,
    \linebreak Max Chromosomes per Breeding $C$}
  \Output{Paraphrase Prompts $\mathbf{P}_\textit{p\_mod}$,
    \linebreak Optimized Paraphrase Prompts $\mathbf{P}_\textit{p}$}
  \SetKwFunction{Add}{Add}
  
  $\mathbf{P}_\textit{p} \gets \textit{Mutate}(\mathbf{P}_\textit{p}) $\;
  \For{$P_\textit{m}$ $\mathbf{in}$ $\mathbf{P}_\textit{pop}$}
  {
    $\mathbf{P}_\textit{p-pop} +\kern-3pt=  \textrm{SentenceParaphrase}(P_\textit{m}, \mathbf{P}_\textit{p}, C)$\;
    $\mathbf{P}_\textit{p-pop} +\kern-3pt=  \textrm{WordParaphrase}(P_\textit{m}, C)$\;
  } 
  \SetKwFunction{Add}{Add}
  
  \Return $\mathbf{P}_\textit{p-pop}$, $\mathbf{P}_\textit{p}$\;
  \caption{Paraphrase}
  \label{algorithm:paraphrase}
\end{algorithm}
 
\subsection{Text Generation}
\label{sub: text_generation}

We generate text for each \textit{task-specific} prompt (e.g.,
\textit{``Text that expresses \emoclass''}) in
$\mathbf{P}_\textit{pop}$ using a large pre-trained language model,
such as GPT-3.5 \cite{openai2022gpt35}, Llama-2
\citep{touvron2023llama}, or Mistral 7B \cite{jiang2023mistral}. To
do this, \emoclass is replaced with each relevant emotion
category -- anger, disgust, fear, joy, or sadness. We refer to these
instantiations as \textit{Text Generation Prompts}.
 
\subsection{Fitness Evaluation}
\label{sub:fitness_evaluation}
Each \textit{task-specific} prompt is evaluated through the texts
generated from its corresponding \textit{Text Generation Prompt}.
The evaluation compares
the intended emotional condition of the prompt with the predictions
made by objective classifiers. The
probability scores for the correct class are used as the objective
value during optimization and final
evaluation. These probability scores are obtained from
two independent classifiers, each trained on separate data.  In the
evaluation, we filter out generated texts that are a paraphrase
of the \textit{Text Generation Prompt}\footnote{We filter out
  texts with a BLEU score > 0.2. For example,
  a language model generates ``The text expresses joy.''
  from the \textit{Text Generation Prompt}: \textit{``Write a text that expresses joy''}.}.

\subsection{Pareto Selection}
\label{sub:pareto_selection}
We utilize the NSGA-II algorithm to rank prompts from the set
$\mathbf{T}_\textit{eval}$, which forms the Pareto front -- the set
of optimal solutions balancing multiple conflicting objectives. While
the ideal in natural language generation is to find a single solution
that maximizes all objectives, this is rarely achievable in
practice. Pareto selection provides a practical approach, allowing us
to identify a set of solutions that represent the best possible
trade-offs between competing objectives.

The NSGA-II uses non-dominated sorting to rank prompts based on their
performance across the objective front. A prompt \( a \) is
non-dominated if no other prompt \( b \) exists such that
\( \forall i, f_i(b) \geq f_i(a) \) and
\( \exists j, f_j(b) > f_j(a) \), where \( f_i \) represents the
objective functions. This approach finds solutions that may not be
perfect for every objective, but are optimal given the inherent
trade-offs.

In addition to the top-n solutions ranked by NSGA-II, we also include
the top-n performing solutions from each individual objective that
were excluded from the Pareto ranking. This inclusion is based on the
assumption that highly objective-specific solutions can contribute
valuable features to the next generation, particularly during genetic
operations such as combination.

\begin{table}[t]
\setlength{\tabcolsep}{4pt}
\centering\footnotesize
\begin{tabularx}{\columnwidth}{lXcccc}
\toprule
\textbf{LLM} & \textbf{Prompt} & \rotatebox{90}{\textbf{ISEAR}} & \rotatebox{90}{\textbf{TEC}} & \rotatebox{90}{\textbf{AT}} & \rotatebox{90}{\textbf{Avg.}}\\
\midrule
Seed &  Write a text that expresses \emoclass & .92 & .60 & .31 & .63 \\
\midrule
GPT-3.5 & I came across \emoclass while $\langle$circumstance$\rangle$ because $\langle$reason$\rangle$. & .99 & .97 & .96 & .97\\
\cmidrule(r){2-6}
Llama & ? Sure! Here's a sentence that combines the key elements of ``The aroma of fresh baked croissants wafted'', ``The rhythmic beats of \emoclass music played in the backgroun'', and "The soothing melodies of the $\langle$class$\rangle$ genre transported me to & .99 & .97 & .94 & .96\\
\cmidrule(r){2-6}
Mistral & Unlock the true potential of \emoclass to craft a compelling and moving expression that resonates deeply with your audience and leaves a profound impact & .99 & .97 & .91 & .95 \\
\bottomrule
\end{tabularx}
\caption{Performance of the best seed prompt and multi-objective optimized prompts for three LLMs. ISEAR, TEC, and Affective Text (AT) columns show their respective fitness evaluations and Average (Avg.) represents the fitness averaged across all objectives.
}

\label{table:prompts_optimization_llms}
\end{table}

\section{Experiments}
We evaluate the Multi-Objective Prompt Optimization (MOPO) algorithm
for affect-driven text generation using three datasets. Each of them
exhibits distinct emotional characteristics. We compare
MOPO to the single-objective method by
\citet{resendiz-klinger-2023-emotion} which is the only approach we
are aware of that studied prompt optimization for text generation (see
Section~\ref{sec:relatedwork}).

\paragraph{Objective Functions.} We use three emotion datasets to
train the emotion classifiers. The ISEAR dataset contains personal
narratives from people across various cultures, capturing emotional
experiences \citep{scherer1994evidence}. AffectiveText includes news
headlines annotated for emotional content and valence
\citep{strapparava-mihalcea-2007-semeval}. TEC is a collection of
tweets labeled with emotions, representing the spontaneous expression
of feelings on social media \citep{mohammad-2012-emotional}. See
Appendix \ref{sec:objective_classfiers_appendix} for more information
on the training and performance of these classifiers.

\paragraph{Language Model.} We employ GPT-3.5\footnote{The total cost
  of the experiments was 80.95 USD. They have been performed in
  April 2024.}, LLama-7B-Chat, and Mistral-7B as the underlying
language models for conditional text generation, paraphrasing, and
crossover operations\footnote{We generate 5 sentences per \textit{Text
    Generation Prompt}. Crossover and Paraphrase generate 3 prompts
  each.}.

\paragraph{Seed Prompts.} We use 10 \textit{task-specific} seed
prompts ($\mathbf{P}\textit{pop}$), as listed in Table
\ref{tab:seed_prompts} in the Appendix. We designed these prompts
based on simplicity and data set specificity. The Combination Prompts ($\mathbf{P}\textit{c}$,
\textit{Mix the two prompts: ``[prompt\_1]'' ``[prompt\_2]'' Into a
  new single sentence.} ), Paraphrase Prompts ($\mathbf{P}\textit{p}$,
\textit{Paraphrase the following sentence into a new sentence:
  ``[prompt]''}), and Fixed Paraphrase Prompts
($\mathbf{P}_{\textit{fix}}$, \textit{Reorganize the sentence to
  convey the same meaning: ``[prompt]''})  were designed following similar strategies. The full list of prompts is
provided in Appendix \ref{sec:seed_prompts}.

\begin{figure}
    \includegraphics[width=\columnwidth]{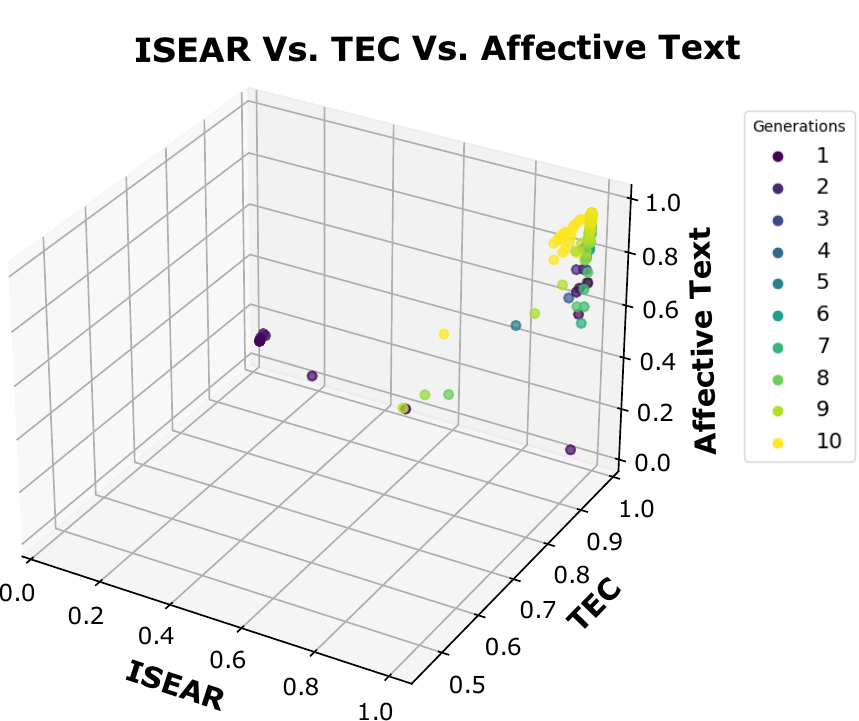} 
    \caption{Improvement in the 10 best-performing prompts from
      Generation 1 (dark blue) to 10 (yellow). Most prompts reach almost a score of 1.}
    \label{fig:3_objectives}
\end{figure}

\begin{table*}[t]
\centering\footnotesize
\setlength{\tabcolsep}{5.5pt}
\begin{tabularx}{\textwidth}{lp{3cm}Xlp{4.5cm}}
\toprule
\textbf{G.} & \textbf{Prompt (Layer-1)} & \textbf{Operation Prompt (Layer-2)} & \textbf{Op.} & \textbf{Generated Text} \\
\midrule
0 & Please complete the sentence: I felt \emoclass when/because & --- & --- & accomplished something I have been working towards for a long time.\\
\midrule
  1 & Finish the sentence by writing: I felt \emoclass due to or when. & \textbf{Please paraphrase the following sentence in a clear and concise manner: ``}Please complete the sentence: I felt \emoclass when/because\textbf{''} & p. & I felt joy when I saw my newborn baby for the first time. \\
\midrule
  2 & Complete the sentence by adding: I experienced \emoclass because or when. & \textbf{Please paraphrase the following sentence in a clear and concise manner: ``}Finish the sentence by writing: I felt \emoclass due to or when.\textbf{''} & p. & I experienced joy because I achieved my goal of running a marathon.\\
\midrule
  3 & Enhance the sentence by incorporating: I encountered \emoclass at the moment when... & \textbf{Rewrite ``}Complete the sentence by adding: I experienced \emoclass because or when.\textbf{'' in a creative and engaging manner while keeping the original message intact.} & p. & I encountered joy at the moment when I saw my loved ones after being apart for so long. \\
\midrule
  4 & I encountered \emoclass at the moment when <reason> due to or while <circumstance>. & \textbf{Carefully examine both ``}Enhance the sentence by incorporating: I encountered \emoclass at the moment when...\textbf{'' and ``}Finish the sentence by stating: I encountered \emoclass due to or while.\textbf{'' before combining their main ideas or themes into a single, coherent sentence incorporating elements from both original statements.} & c. & I encountered joy at the moment when my team won the championship game due to our hard work and dedication throughout the season. \\
\bottomrule
\end{tabularx}
\caption{Example of prompt optimization over four generations (G.), where generation 0 is the seed prompt. The Operation Prompt column shows the \textbf{genetic operation prompt} used to improve the Layer-1 prompt from the previous generation. The Operation (Op.) column specifies the genetic operation: paraphrase (p.) or combine (c.). All generated texts are for the emotion (\emoclass) Joy.}
\label{table:prompt_conversion}
\end{table*}

\paragraph{Single- \& Multi-objective optimization.}
We start the prompt optimization process with the same set of seed prompts over 10 generations across four setups: (1) Multi-objective optimization applying three objective functions simultaneously, and (2--4) Single-objective, using each objective individually.

\paragraph{Single-Objective Baseline.} Similar to MOPO, we use the
same objective functions (classifiers), seed prompts, and language models for the
single-objective automatic prompt optimization.

\section{Results}
\subsection{RQ1: Multi-objective vs.\ single-objective optimization}
\label{sub: single-objective}

We begin by evaluating the generalization performance of
multi-objective optimization. We compare multi- and single-objective
optimized prompts against seed prompts to confirm that the
process generally works. Then, we compare
multi- vs.\ single-objectively optimized prompts.

\paragraph{Multi-objective.}
Table \ref{table:prompts_optimization_llms} 
compares seed prompts with optimized prompts using three different LLMs. MOPO improves the macro-average score by up to 34 pp (GPT-3.5) and by at least 25 pp (Mistral). We focus on GPT-3.5 because it outperforms Llama-7B and Mistral-7B. The consistent high fitness scores -- .99 (ISEAR), .97 (TEC), and .96
(Affective Text) -- demonstrate effective multi-objective
optimization. Corresponding results and analyses for the other models are available in the appendix.

\begin{figure*}
    \includegraphics[width=\textwidth]{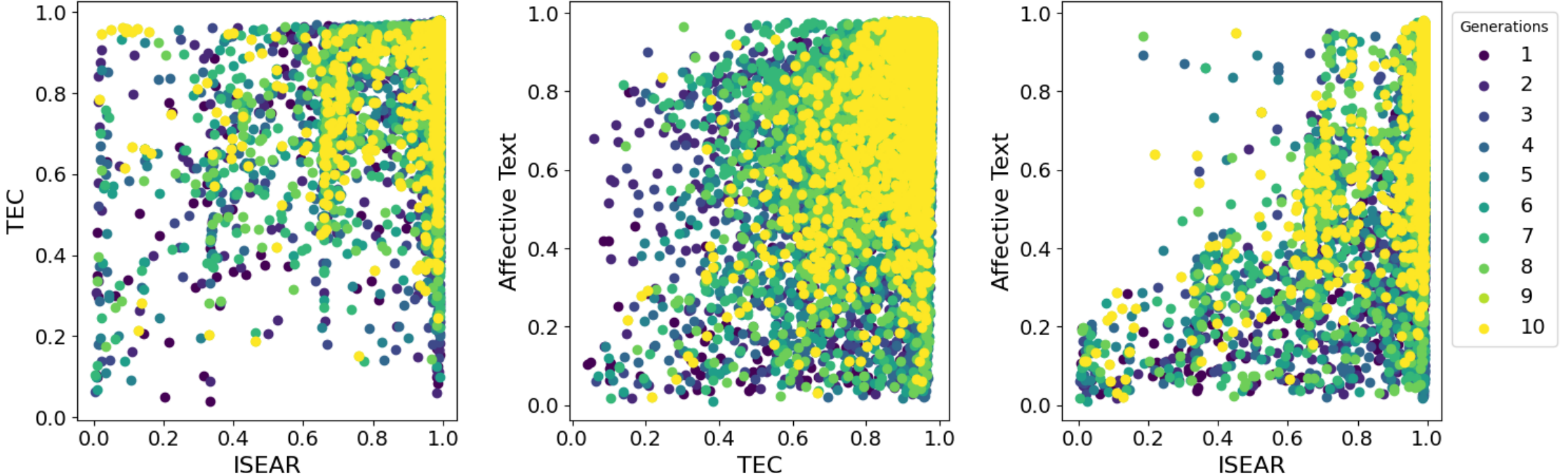} 
    \caption{Improvement across generations of the best-performing prompts for the emotion \textit{joy}. Comparing two objectives at the time. In the last generation (yellow) most of the prompts are close to 1 score (optimal performance).}
    \label{fig:joy_optimization}
\end{figure*}

Table \ref{table:prompt_conversion} traces the operations in the
optimization process of the best prompts. It shows examples of
generated text across generations.  Figure~\ref{fig:3_objectives}
shows the optimization process across all emotions, while
Figure~\ref{fig:joy_optimization} focuses on the emotion \textit{joy}
-- comparing two of the three objectives simultaneously. Optimization
results for all emotions are provided in
Figure~\ref{fig:3_ech_emotion} in the appendix. Both plots demonstrate
a successful optimization process: initially, prompts (darker colors)
have lower fitness, but as optimization progresses, the final
generation (yellow) achieves high fitness across all objectives.

Finally, Table~\ref{table:prompts_optimization_layer_2} presents a
sample of the (self-optimized) Layer-2 prompts that generated the
best-performing Layer-1 prompts for GPT-3.5 (Table
\ref{table:prompts_optimization_llms}), during the final
generation. Appendix~\ref{sec:text_examples} provides the complete set
of optimized prompts derived from the seed layer prompts. Compared to
the seed Layer-2 prompts (Tables~\ref{tab:paraphrasing_prompts} and
\ref{tab:crossover_prompts} in the appendix), the optimized prompts
are more specific and descriptive.
\begin{table}[t]
\setlength{\tabcolsep}{6pt}
\centering\footnotesize
\begin{tabularx}{\columnwidth}{lXp{3cm}}
\toprule
\textbf{Op.} & \textbf{Layer-3 Prompt (Fix)} & \textbf{G. Layer-2 Prompt} \\
\midrule
p.
& \textbf{Rephrase the sentence by changing the form of the words: ``}Paraphrase the  following sentence into a new sentence:
``SENTENCE\_1''\textbf{''}
& Transform the following sentence into a different sentence: ``SENTENCE\_1''\\
\midrule
c.
& \textbf{Paraphrase the following sentence: "}Combine ``SENTENCE\_1'' and ``SENTENCE\_2'' to create a new, cohesive sentence that retains elements from both.\textbf{''}
& Merge ``SENTENCE\_1'' and ``SENTENCE\_2'' to form a fresh, unified statement that incorporates aspects of both.\\
\bottomrule
\end{tabularx}
\caption{Example of the final optimization process for Layer-2 prompts using \textbf{Layer-3 Prompts} (fix). The Operation (Op.) column specifies the genetic operation: paraphrase (p.) or combine (c.), from the final generation. ``SENTENCE\_1'' and ``SENTENCE\_2''are place holder for a Layer-1 prompt.}
\label{table:prompts_optimization_layer_2}
\end{table}
\begin{table*}[t]
\setlength{\tabcolsep}{4pt}
\centering\footnotesize
\begin{tabularx}{\textwidth}{lXccccc}
\toprule
\textbf{O.} & \textbf{Prompt} & \textbf{Opt.} & \rotatebox{90}{\textbf{ISEAR}} & \rotatebox{90}{\textbf{TEC}} & \rotatebox{90}{\textbf{AT}} & \rotatebox{90}{\textbf{Avg.}}\\
\midrule
\multirow{6}{*}[-4em]{\rotatebox{90}{Single}} & In formal writing, finish the sentence with ``I experienced \emoclass emotions when / due to.'' In informal writing, finish it with ``I felt <class> feelings when / due to. & ISEAR & .99 & .93 & .74 & .88\\
 & Complete the statement: 1. He experienced \emoclass as a result of <reason>. & ISEAR & .99 & .90 & .74 & 87\\
  \cmidrule{2-7}
 & When I think about the defining essence of \emoclass, it shines unconditionally at its core, especially in <specific situation>, where <class> shines brightly during moments of <description>. This display embodies the purest form of \emoclass and leaves a lasting impact on all who witness it. & TEC & .98 & .97 & .74 & .89 \\
 & The essence of \emoclass is truly illuminated in <specific situation>, embodying \emoclass in a compelling and impactful manner. & TEC & .98 & .97 & .75 & .90\\
\cmidrule{2-7}
 & Certainly! The request is for someone to send a text message stating, ``I feel prepared and confident to rock \emoclass!'' & AT & .98 & .95 & .98 & .97 \\
 & Please send me a text saying 'I feel prepared and confident to rock \emoclass!' & AT & .98 & .96 & .98 & 97 \\
\midrule
\multirow{2}{*}[-.5em]{\rotatebox{90}{Multi}} & I came across \emoclass while $\langle$circumstance$\rangle$ because $\langle$reason$\rangle$. & All & .99 & .97 & .96 & .97\\
 & How does the powerful language of \emoclass affect individuals deeply involved in it, and have you witnessed someone being deeply touched by words that perfectly captured their experience in \emoclass? & All & .98 & .97 & .96 & .97\\
\bottomrule
\end{tabularx}
\caption{Performance of the two top-performing single- and
  multi-objective optimized prompts. The Optimization (Opt.) column
  shows the objective used for optimization: ISEAR, TEC, or Affective
  Text (AT) for single-objective, and All for multi-objective. The
  ISEAR, TEC, and AT columns indicate the fitness scores for each
  respective objective. The Average (Avg.) column represents the
  averaged score across all objectives.}
\label{table:prompts_optimization}
\end{table*}

\paragraph{Single-objective.} Table \ref{table:prompts_optimization} presents scores for the best-performing single- and multi-objective optimized prompts, and the optimization objective (Opt.) used. Optimizing for a specific objective improves its performance notably more than for others -- diagonal scores are higher under the single-objective (O.) section. However, these optimizations also expose generalization challenges across datasets: ISEAR and TEC prompts achieve high mutual scores (above .90, columns ISEAR and TEC) but fall short in matching the style of AffectiveText when evaluated outside their optimization context (Rows 1--4). In contrast, prompts optimized for AffectiveText demonstrate a higher ability to produce text resembling ISEAR and TEC content (Rows 5,7). This implies that news headlines are more challenging to classify, which often imply emotions indirectly (e.g., ``UK announces immigration restrictions'') compared to the explicit emotional expressions in self-reports or tweets (e.g., ``I feel happy  \#WatchingTheSunset''), from the ISEAR and TEC datasets.

\paragraph{Single- vs.\ Multi-objective.} 
We now want to understand if multi-objective optimization comes with a
loss or gain in single-objective values, when optimized only for
them. Table~\ref{table:single_vs_multiple} compares the performance of
single-objective (S.\ Obj columns) with multi-objective (M.\ Obj) and
the difference (M.\ vs.\ S.) across the three objectives
(rows). Multi-objective prompts perform similarly to the best
individual single-objective prompts, with only a small loss for AT (2
pp, diagonal in M.\ vs.\ S.). However, the best multi-objective
prompts can achieve noticeable improvements in other domains (up to 6
pp for TEC and up to 25 pp for AT), suggesting that multi-objective
optimization enhances generalizability across different
datasets. These findings indicate that while single-objective
optimization may be sufficient for specific tasks, multi-objective
optimization can provide broader benefits across various domains.%

\begin{table}[t]
\centering \footnotesize
\begin{tabularx}{\columnwidth}{Xccc|c|rrr}
\toprule
 & \multicolumn{3}{c}{\textbf{S. Obj}} & \multicolumn{1}{c}{\textbf{M.}} & \multicolumn{3}{c}{\textbf{M. vs. S.}} \\
 \cmidrule(lr){2-4} \cmidrule(lr){6-8}
 & \rotatebox{90}{\textbf{ISEAR}} & \rotatebox{90}{\textbf{TEC}} & \rotatebox{90}{\textbf{AT}} & \textbf{Obj.} & \rotatebox{90}{\textbf{ISEAR}} & \rotatebox{90}{\textbf{TEC}} & \rotatebox{90}{\textbf{AT}} \\
\midrule
ISEAR & .99 & .93 & .74 & .99 & 0 & $+$.06 & $+$.25 \\
TEC & .98 & .97 & .74 & .97 & $-$.01 & 0 & $+$.23 \\
AT & .98 & .95 & .98 & .96 & $-$.02 & $+$.01 & $-$.02 \\
\midrule
Avg. & .98 & .95 & .82 & .97 & $-$.01 & $+$.02 & $+$.15 \\
\bottomrule
\end{tabularx}
\caption{Comparison between Single-Objective (S. Obj.) and Multi-Objective (M. Obj.) prompt optimization. ISEAR, TEC, and Affective Text (AT) rows show evaluations from the best-performing prompt. The M. vs. S. columns indicate the improvement or decrease of Multi-objective optimization compared to Single-objective.}
\label{table:single_vs_multiple}
\end{table}
\subsection{RQ2: How do paraphrasing and combining prompts affect performance?}
To understand if both paraphrasing prompts and combining them have an
impact on the overall optimization performance, we individually remove
the operations to evaluate their impact, using the same objectives and
seed prompts as the multi-objective optimization in Section \ref{sub:
  single-objective}.  Table \ref{table:ablation} shows the results of
this ablation study.  The results reveal that removing
\textit{Combination} decreases performance by 4 pp, and omitting
\textit{Paraphrase} by 1 pp on average across all objectives. These
findings are consistent with their contribution to generating the
top-n prompts in each generation. Paraphrase generate 88\% of the
prompts in the Pareto front, and Combination 12\%.

\subsection{RQ3: Does objective optimization impact the quality of the generated text?}
To understand if the optimization paradigm impacts the language
quality, we perform an automatic and a human annotation study.  We use
GPT-3.5, known for its ability to match human performance in text
quality assessment \cite{chiang-lee-2023-closer, liu-etal-2023-g}, and
three human annotators. The evaluation focuses on Coherence, Fluency,
Grammar, Plausibility, Native Speaker Likeness, and Human Likeness. We
adopt a 5-point Likert scale, ranging from 1 (strongly disagree) to 5
(strongly agree), to rate each dimension of text quality (Table
\ref{tab:text_quality_prompts}).
\begin{table}[t]
\centering\small
\begin{tabular}{lrrrr}
\toprule
\textbf{Config.} & \textbf{ISEAR} & \textbf{TEC} & \textbf{AT} &
                                                                 \textbf{Avg.}\\
  \midrule
All & .99 & .96 & .94 & \textbf{.96} \\
No Combination & .99 & .96 & .81 & .92\\ 
No Paraphrase & .99 & .95 & .92 & .95\\
 \bottomrule
\end{tabular}
\caption{Ablation study for MOPO's genetic operations using the ten best-performing prompts.}
\label{table:ablation}
\end{table}
\begin{table}[t]
\centering\small
\setlength{\tabcolsep}{5pt}
\begin{tabular}{llcccccc}
\toprule
\rotatebox{90}{\textbf{Evaluation}} & \rotatebox{90}{\textbf{Dataset}}
  & \rotatebox{90}{\textbf{Fluency}} & \rotatebox{90}{\textbf{Native
                                       Spkr}} &
                                                \rotatebox{90}{\textbf{Coherency}} & \rotatebox{90}{\textbf{Plausability}} & \rotatebox{90}{\textbf{W. by AI}} & \rotatebox{90}{\textbf{W. by human}} \\
\midrule
\multirow{7}{*}{\rotatebox{90}{GPT-3.5}} & MOPO-All & 4.1 & 4.1 & 3.7 & 3.0 & 3.8 & 4.5 \\
& MOPO-ISEAR & 3.9 & 4.3 & 3.5 & 3.8 & 3.8 & 4.7 \\ 
& MOPO-Tec & 4.1 & 4.1 & 4.0 & 3.5 & 3.8 & 4.4 \\
& MOPO-AT & 3.9 & 3.6 & 3.1 & 2.9 & 3.8 & 4.4 \\\cmidrule{2-8}
& ISEAR & 3.0 & 3.0 & 2.1 & 2.8 & 3.9 & 4.1 \\
& TEC & 3.2 & 3.1 & 2.2 & 2.6 & 3.7 & 3.9 \\
& AT & 4.1 & 4.4 & 3.0 & 3.1 & 3.8 & 4.5 \\\midrule
H. & MOPO-All & 3.4 & 3.1 & 2.9 & 2.4 & 3.9 & 3.5 \\
 \bottomrule
\end{tabular}
\caption{Text quality evaluation was conducted using both GPT-3.5 and
  human evaluators (H.) on a five-point Likert scale, where 1 means
  ``strongly disagree'' and 5 means ``strongly agree'' (higher is
  better).}
\label{table:gramma}
\end{table}
For the automatic evaluation, we randomly sampled 1,000 texts from the
final outputs of MOPO, covering both single- and multi-objective
setups, as well as from the ISEAR, TEC, and AffectiveText
datasets. For the human evaluation, we sample 100 instances from the
multi-objective optimization (MOPO-All).

Table \ref{table:gramma} shows the results. Generally, the
AffectiveText dataset yields higher scores, closely followed by
MOPO-generated texts. This discrepancy may stem from AffectiveText's
professionally written and reviewed headlines. Nonetheless, the
majority of scores fall within an acceptable range. Text quality is
largely influenced by the language model itself rather than the
optimization objective(s) -- MOPO's generated texts maintain similar
quality across different objectives (see Appendix \ref{sec:Text Quality appendix} for an analysis of MOPO with the other LLMs). However, the model conditioned on
Affective Text produces lower-quality text compared to other
configurations, implying that generating headlines is
challenging. This may account for the low scores observed in Section
\ref{sub: single-objective}.

\subsection{State-of-the-art Baseline}
Table \ref{table:sota} compares SOTA (top) prompt optimization with
MOPO (bottom) using three LLMs as base models. MOPO outperform the
SOTA optimizations across all objectives. Similar to Section \ref{sub:
  single-objective}, SOTA for a single objective struggles to
generalize across objectives. The underlying LLMs show similar
performance trends, with GPT-3.5 outperforming Llama2 and
Mistral. These results demonstrate MOPO's superiority over SOTA
methods for prompt optimization. Additionally, MOPO allows users to
select the best prompt for a specific objective or one that
generalizes across all objectives -- no multiple optimizations are
required.

\begin{table}[t]
\centering \footnotesize
\begin{tabularx}{\columnwidth}{cXcccc}
\toprule

 & \textbf{Model} & \textbf{ISEAR} & \textbf{TEC} & \textbf{AT} & \textbf{Avg.} \\
 \midrule
 
\multirow{9}{*}{\rotatebox{90}{SOTA}} & Llama2-ISEAR & .99 & .92 & .49 &  .80 \\
&  Llama2-TEC & .98 & .97 & .55 & .83  \\
&  Llama2-AT & .96 & .94 & .60 &  .83 \\
\cmidrule{2-6}

& Mistral-ISEAR & .99 & .95 & .46 & .80  \\
& Mistral-TEC & .99 & .97 & .57 & .84 \\
& Mistral-AT & .98 & .95 & .63 &  .85 \\
\cmidrule{2-6}

& GPT-3.5-ISEAR & .99 & .90 & .83 & .90 \\
& GPT-3.5-TEC & .94 & .97 & .70 & .87 \\
& GPT-3.5-AT & .97 & .91 & .88 & .92 \\
\midrule

\multirow{3}{*}{\rotatebox{90}{MOPO}} & GPT-3.5-All & .99 & .97 & .96 & .97  \\

& Llama2-All & .99 & .97 & .94 & .96 \\
& Mistral-All & .99 & .97 & .69 & .88  \\

\bottomrule
\end{tabularx}
\caption{Comparison between state-of-the-art prompt optimization \citep{resendiz-klinger-2023-emotion} and MOPO. ISEAR, TEC, and Affective Text (AT) rows show evaluations from the best-performing prompt.}
\label{table:sota}
\end{table}

\section{Conclusion and Future Work}
In this paper, we have shown the first algorithm that optimizes
prompts multiobjectively. We see that the performance increases
substantially across multiple objectives -- which single-objective
optimization cannot achieve -- with only a minimal loss (1--2
pp). Additionally, MOPO eliminates the need for separate optimizations
for each objective. MOPO uses a self-referential process to optimize
task-specific and mutation/combination prompts.

This leads to important future work. We focused on affective text
generation, but MOPO's design is generic. Therefore, we suggest to
evaluate it across various setups, including machine translation,
question-answering, and text classification. Investigating the
limitations concerning the number of objectives, such as optimizing a
single prompt for multiple languages or LLM models, is
crucial. Additionally, our current method treats combination and
mutation equally. Alternative approaches to learning in the Markov
decision process, like reinforcement learning, could offer more
efficient prompt selection and variation strategies.

\section*{Acknowledgements}
This work has been supported by a CONACYT scholarship
(2020-000009-01EXTF-00195) and by the German Research Council (DFG),
project ``Interactive Prompt Optimization with the Human in the Loop
for Natural Language Understanding Model Development and
Intervention'' (\textsc{InPrompt}, KL 2869/13-1, project number
521755488).
 
\section*{Ethical Considerations}
\label{sec:ethical_considerations}
The proposed methodology aims to optimize prompts with one or more
objectives, but MOPO must be used cautiously to avoid risks. Optimized
prompts might produce harmful content, such as discriminatory
language, misinformation, fake news, or imitations of specific
individuals or groups, if such conditions are set as
objectives. Therefore, responsible and ethical use of MOPO is
essential.

Additionally, the underlying risks associated with the base
pre-trained language models (e.g., GPT, Llama-2, FLAN) must be
considered. These models may have been trained on biased data,
potentially leading to text that perpetuates stereotypes or
marginalizes certain groups. It is important to note that such risks
are not inherent to the MOPO methodology but stem from the base models
used.

\section*{Limitations}

The effectiveness of our proposed method largely depends on the base
language models (e.g., GPT, LLama-7B-Chat, and Mistral-7B) used to
modify and combine the initial seed prompts. The number of generations
needed can vary significantly depending on the underlying model used
and genetic operations. Additionally, the objective functions are
crucial as they direct the entire optimization process, and they can
be sensitive to their initial setup, tuning, and performance.

There are several limitations to consider in each module of our
approach. First, the variability of outcomes based on the choice of
the base language model means that different models may require
varying numbers of generations to achieve optimal results. Second,
while the genetic operations facilitate diversity in prompt
generation, they can introduce unpredictability in performance across
different tasks. Third, the number of samples generated from the
genetic operations ($\mathbf{P}\textit{c}$ and $\mathbf{P}\textit{p}$)
and the \textit{Text Generation Prompt} may influence the convergence
of the objectives. Fourth, the objective functions themselves may not
fully capture the complexity of the task, potentially leading to less
optimal results in some cases.

Another important limitation is that each run of the experiment setup
was conducted only once, meaning that the results may not account for
variability or potential improvements that could arise from multiple
iterations.

Overall, this method has proven useful for generating text based on
specific emotions, it is important for users to be aware of these
limitations when considering its capabilities and applications. We
encourage users to keep these limitations in mind when evaluating the
method's capabilities and applications.

\bibliography{anthology.bib}

\newpage
\clearpage
\appendix

\section{Prompts}
\label{sec:seed_prompts}

We utilized 10 Layer-1 prompts for conditional text generation, as
shown in Table \ref{tab:seed_prompts}. These prompts were chosen for
their simplicity and include two questions taken directly from the
ISEAR dataset. For Layer-2, we applied similar techniques to create
Paraphrase Prompts (Table \ref{tab:paraphrasing_prompts}) and
Crossover Prompts (Table \ref{tab:crossover_prompts}). Lastly, the
Layer-3 unoptimized prompts, which only mutate Layer-2 prompts, are
detailed in Table \ref{tab:fix_prompts}.

\begin{table}[h]
\centering \footnotesize
\begin{tabularx}{\columnwidth}{X}
\toprule
\textbf{Layer-1: Text Generation Prompt} \\
\midrule
Describe a situation where a person felt \emoclass \\
Write a text that expresses \emoclass \\
Phrases that express \emoclass \\
What is a sentence example for \emoclass ? \\
Can you provide an example of a situation where someone experienced \emoclass? \\
What is an example of a \emoclass sentence? \\
\emoclass sentence \\
Experience for \emoclass? \\
Please describe a situation or event — in as much detail as possible — in which a reader felt \emoclass \\
Please complete the sentence: I felt \emoclass when/because \\
\bottomrule
\caption{List of Seed Prompts for Conditional Text Generation: During generation, each prompt is replicated across all emotions, substituting the \emoclass token with the respective emotion.}
\label{tab:seed_prompts}
\end{tabularx}
\end{table}

\begin{table}[h]
\centering \footnotesize
\begin{tabularx}{\columnwidth}{X}
\toprule
\textbf{Layer-2: Paraphrase Prompt} \\
\midrule
Paraphrase the following sentence into a new sentence: ``SENTENCE\_1'' \\
Given the following sentence: ``SENTENCE\_1'' Paraphrase the sentence into a new one by keeping the same meaning. \\
Please paraphrase the following sentence in a clear and concise manner: ``SENTENCE\_1'' \\
Rewrite ``SENTENCE\_1'' in a more formal (or informal) tone while retaining the original meaning. \\
Simplify ``SENTENCE\_1'' for a younger audience without changing its meaning. \\
Expand ``SENTENCE\_1'' into a more detailed explanation without altering its original intent. \\
Creatively rewrite ``SENTENCE\_1'', ensuring the new version is engaging yet maintains the same message. \\
Summarize ``SENTENCE\_1'' in fewer words, ensuring the main idea is fully intact \\
Rewrite ``SENTENCE\_1'' from a different perspective (e.g., first person to third person), keeping the essence the same. \\
Can you simplify this sentence to make it easier to understand? ``SENTENCE\_1'' \\
\bottomrule
\end{tabularx}
\caption{List of Paraphrase Prompts (Layer-2): In each generation, ``SENTENCE\_1'' is substituted with a Layer-1 prompt.}
\label{tab:paraphrasing_prompts}
\end{table}

\begin{table}[h]
\centering \footnotesize
\begin{tabularx}{\columnwidth}{X}
\toprule
\textbf{Layer-3: Paraphrase Prompt} \\
\midrule
Reorganize the sentence to convey the same meaning: ``SENTENCE\_1''\\
Transform the sentence to a different voice or perspective: ``SENTENCE\_1'' \\
Paraphrase the following sentence: ``SENTENCE\_1'' ''\\
Rewrite the sentence using different words: ``SENTENCE\_1''\\
Paraphrase the sentence with a more casual tone: ``SENTENCE\_1''\\
Rephrase the sentence by changing the form of the words: ``SENTENCE\_1''\\
\bottomrule
\end{tabularx}
\caption{List of Level-3 Fixed (Unoptimized) paraphrase Seed Prompts, exclusively optimizing Crossover (Pc) and Paraphrase (Pp) prompts. ``SENTENCE\_1'' is substituted with either a crossover or paraphrase prompt at each generation of the optimization.}
\label{tab:fix_prompts}
\end{table}

\begin{table*}
\centering \footnotesize
\begin{tabularx}{\textwidth}{X}
\toprule
\textbf{Layer-2: Crossover Prompt} \\
\midrule
The following two sentences are prompts for conditional text generation. ``SENTENCE\_1''``SENTENCE\_2'' Summarize both prompts into one. \\
Mix the two prompts: ``SENTENCE\_1'' ``SENTENCE\_2'' Into a new single sentence. \\
Combine ``SENTENCE\_1'' and ``SENTENCE\_2'' to create a new, cohesive sentence that retains elements from both. \\
Merge the themes of ``SENTENCE\_1'' and ``SENTENCE\_2'' into a single sentence that seamlessly integrates their ideas. \\
Craft a new sentence by blending the key elements of ``SENTENCE\_1'' and ``SENTENCE\_2'', ensuring that the final sentence is coherent and flows naturally. \\
Formulate a new sentence that synthesizes the concepts from ``SENTENCE\_1'' and ``SENTENCE\_2'', maintaining a balance between the two. \\
Create a cohesive and fluent sentence that intertwines the essence of both ``SENTENCE\_1'' and ``SENTENCE\_2''. \\
Read ``SENTENCE\_1'' and ``SENTENCE\_2''. Then, synthesize their main ideas or themes into a new, single sentence. Ensure that the new sentence reflects elements from both original sentences in a balanced and coherent way. \\
Analyze the content and tone of ``SENTENCE\_1'' and ``SENTENCE\_2''. Use this analysis to construct a new sentence that merges the essence of both, maintaining the style and tone present in the original sentences. \\
Identify the key elements or messages in ``SENTENCE\_1'' and ``SENTENCE\_2''. Create a new sentence that weaves these elements together, ensuring the resulting sentence is harmonious and fluid, and preserves the intent of both original sentences. \\
Examine ``SENTENCE\_1'' and ``SENTENCE\_2'' for their unique characteristics. Then, blend these characteristics to produce a new sentence that seamlessly combines the distinct qualities of both into a unified, coherent statement. \\
Consider the context and underlying themes in ``SENTENCE\_1'' and ``SENTENCE\_2''. Use this insight to generate a new sentence that encapsulates the themes or messages from both in a cohesive and eloquent manner. \\
Interpret the imagery or concepts presented in ``SENTENCE\_1'' and ``SENTENCE\_2''. Develop a new sentence that intertwines these images or concepts, ensuring the new sentence is clear, concise, and effectively communicates the blended ideas. \\
Reflect on the narrative or descriptive elements in ``SENTENCE\_1'' and ``SENTENCE\_2''. Fuse these elements into a new sentence that tells a story or paints a picture, combining the narratives or descriptions from both original sentences. \\
\bottomrule
\end{tabularx}
\caption{List of Crossover (Layer-2) Prompts: ``SENTENCE\_1'' and ``SENTENCE\_2'' are replaced with the highest-performing Layer-1 prompts from the previous generation throughout each optimization cycle.}
\label{tab:crossover_prompts}
\end{table*}

\begin{table*}
\centering \footnotesize
\begin{tabularx}{\textwidth}{lX}
\toprule
\textbf{Metric} & \textbf{Prompt} \\
\midrule
Fluency & Assess the text's fluency, assigning a score from 1 to 5, with 5 representing the highest level of fluency. Do not give an explanation of the selection. \\
Native Speaker & Assess whether the text was written by a native English speaker, assigning a score from 1 to 5, with 5 indicating native-level. Do not give an explanation of the selection. \\
Coherence & Assess the text's coherence, assigning a score from 1 to 5, with 5 representing the highest level of coherence. Do not give an explanation of the selection. \\
Plausibility & Assess the plausibility of the events described in the text, assigning a score from 1 to 5, with 5 indicating the highest level of plausibility. Do not give an explanation of the selection. \\
Written by AI & Evaluate the likelihood that the text was written by AI, assigning a score from 1 to 5, with 5 indicating a strong likelihood of AI authorship. Do not give an explanation of the selection. \\
Written by Human & Evaluate the likelihood that the text was written by a human, assigning a score from 1 to 5, with 5 indicating a strong likelihood of human authorship. Do not give an explanation of the selection. \\
\bottomrule
\end{tabularx}
\caption{Text Quality Assessment Criteria prompts to evaluate the quality of generated text($\mathbf{T}_\textit{pop}$), employing a 5-point Likert scale from 1 (strongly disagree) to 5 (strongly agree).}
\label{tab:text_quality_prompts}
\end{table*}

\section{Objective Classifiers}
\label{sec:objective_classfiers_appendix}

We use three emotion datasets to train the classifiers, which will serve as objective functions during the optimization process. Table \ref{table:F1_score_objective_functions} shows the F1 scores over the five subsets of emotions they have in common -- anger, disgust, fear, joy and sadness. The classifiers were trained on top of RoBERTa \cite{liu2019roberta} using standard parameters for 10 epochs on an NVIDIA RTX A6000 GPU. The International Survey on Emotion Antecedents and Reactions (ISEAR) includes personal narratives from individuals across various cultures. The AffectiveText dataset consists of news headlines annotated for emotional content and valence, providing a distinct insight into how emotions are portrayed in the media. The Twitter Emotion Corpus (TEC) is a collection of tweets labeled with emotions, capturing the spontaneous expression of feelings on social media. 

\begin{table}[H]
\centering\small
\setlength{\tabcolsep}{4pt}
\begin{tabular}{lrrrrrr}
\toprule
\textbf{Dataset} & \textbf{Anger} & \textbf{Disgust} & \textbf{Fear}
  & \textbf{Joy}  & \textbf{Sadness} & \textbf{Avg.}\\
  \midrule

ISEAR & 0.78 & 0.80 & 0.91 & 0.96 & 0.87 & 0.86 \\
TEC  & 0.56 & 0.68 & 0.71 & 0.81 & 0.68 & 0.69\\ 
AT & 0.81 & 0.53 & 0.92 & 0.96 & 0.92 & 0.82\\
 \bottomrule
\end{tabular}
\caption{ F1 scores of the ISEAR, TEC, and AffectiveText (AT) classifiers, used as objective functions during MOPO's optimization process.}
\label{table:F1_score_objective_functions}
\end{table}

\section{Pareto Front}
\label{sec:Pareto_front}

\begin{figure*}
    \includegraphics[width=\textwidth]{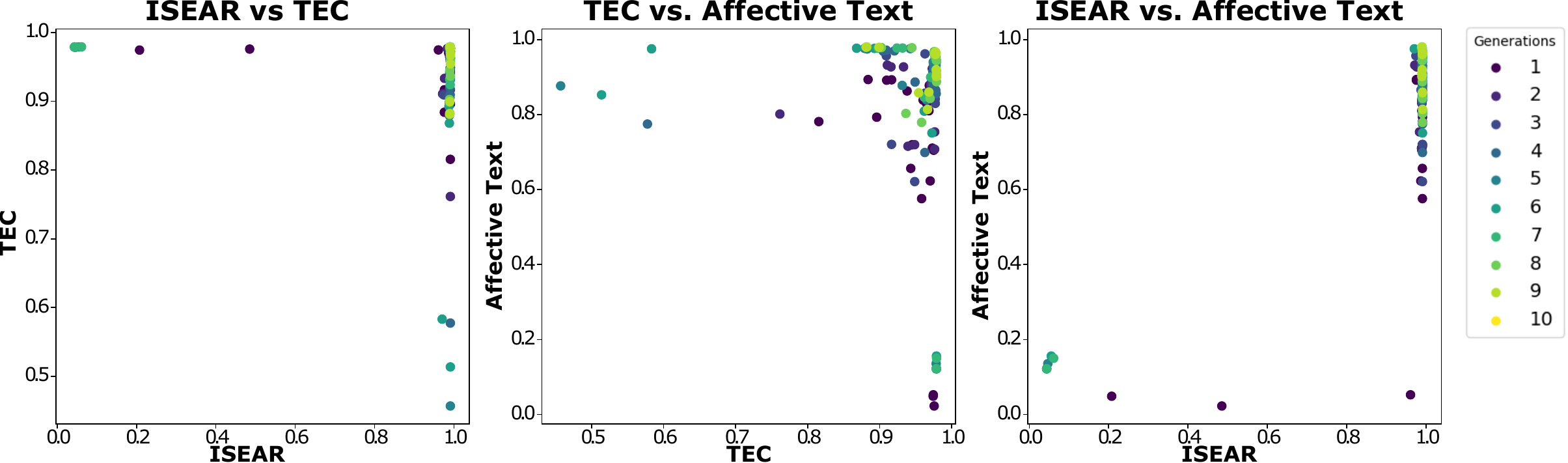} 
    \caption{Improvement across generations of the best-performing prompts, starting in generation 1 to 10. Comparing two objectives at the time. In the last generation, most of the prompts are close to 1 score.}
    \label{fig:3_objectives_pairs}
\end{figure*}

\begin{table*}[t]
\centering\footnotesize
\setlength{\tabcolsep}{2pt}
\begin{tabularx}{\textwidth}{lp{4cm}Xp{4cm}}
\toprule
\textbf{Op.} & \textbf{Layer-2 Prompt} & \textbf{Layer-3 Prompt (Fix)} & \textbf{Generated Layer-2 Prompt} \\
\midrule
p. & Given the following sentence:
``SENTENCE\_1''
Paraphrase the sentence into a new one by keeping the same meaning.
& \textbf{Transform the sentence to a different voice or perspective: ``}Given the following sentence:
``SENTENCE\_1''
Paraphrase the sentence into a new one by keeping the same meaning.\textbf{''}

& The following sentence is given: ``SENTENCE\_1''. Please rephrase it while keeping the original message intact. \\
\cmidrule(r){2-4}

p. & Paraphrase the  following sentence into a new sentence:
``SENTENCE\_1''
& \textbf{Rephrase the sentence by changing the form of the words: ``}Paraphrase the  following sentence into a new sentence:
``SENTENCE\_1''\textbf{''}
& Transform the following sentence into a different sentence: ``SENTENCE\_1''\\
\cmidrule(r){2-4}

p. & Given the following sentence:
``SENTENCE\_1''
Paraphrase the sentence into a new one by keeping the same meaning.
& \textbf{Reorganize the sentence to convey the same meaning: ``}Given the following sentence:
``SENTENCE\_1''
Paraphrase the sentence into a new one by keeping the same meaning.\textbf{''}
& Paraphrase the sentence by keeping the same meaning: ``SENTENCE\_1'' \\

\midrule

c. & Read ``SENTENCE\_1'' and ``SENTENCE\_2''. Then, synthesize their main ideas or themes into a new, single sentence. Ensure that the new sentence reflects elements from both original sentences in a balanced and coherent way.
& \textbf{Reorganize the sentence to convey the same meaning: ``}Read ``SENTENCE\_1'' and ``SENTENCE\_2''. Then, synthesize their main ideas or themes into a new, single sentence. Ensure that the new sentence reflects elements from both original sentences in a balanced and coherent way.\textbf{''}
& Read ``SENTENCE\_1'' and ``SENTENCE\_2'', and synthesize their main ideas or themes into a new, single sentence that effectively combines elements from both."\\
\cmidrule(r){2-4}
c. & Paraphrase the sentence with a more casual tone: ``SENTENCE\_1''
& \textbf{Paraphrase the sentence with a more casual tone: ''}Merge the themes of ``SENTENCE\_1'' and ``SENTENCE\_2'' into a single sentence that seamlessly integrates their ideas.\textbf{``}
& Combine the ideas from ``SENTENCE\_1'' and ``SENTENCE\_2'' into one sentence that flows smoothly.\\
\cmidrule(r){2-4}
c & Paraphrase the following sentence: ``SENTENCE\_1''
& \textbf{Paraphrase the following sentence: "}Combine ``SENTENCE\_1'' and ``SENTENCE\_2'' to create a new, cohesive sentence that retains elements from both.\textbf{''}
& Merge ``SENTENCE\_1'' and ``SENTENCE\_2'' to form a fresh, unified statement that incorporates aspects of both.\\

\bottomrule
\end{tabularx}
\caption{Example of prompt optimization for Layer-2 prompts using Layer-3 (fix). The Operation (Op.) column specifies the genetic operation: paraphrase (p.) or combine (c.). The \textbf{Layer-3 prompt} is used to optimize the Layer-2 prompt, resulting in a new generated Layer-2 prompt. }
\label{tab:layer2 optimization}
\end{table*}

Figure \ref{fig:3_objectives} shows the improvement of top-performing prompts in a three-objective optimization, comparing the TEC, ISEAR, and Affective Text objectives in pairs. The ISEAR dataset shows the highest compatibility with the other two, as shown by the large number of prompts achieving high scores (few dots in the middle of the plots, in the extreme plots). In contrast, the TEC and Affective Text datasets initially exhibit more conflict, with prompt performance starting low. However, as optimization progresses, performance improves, moving towards the upper right corner of the plots.

\begin{figure*}
\center
    \includegraphics[scale=.37]{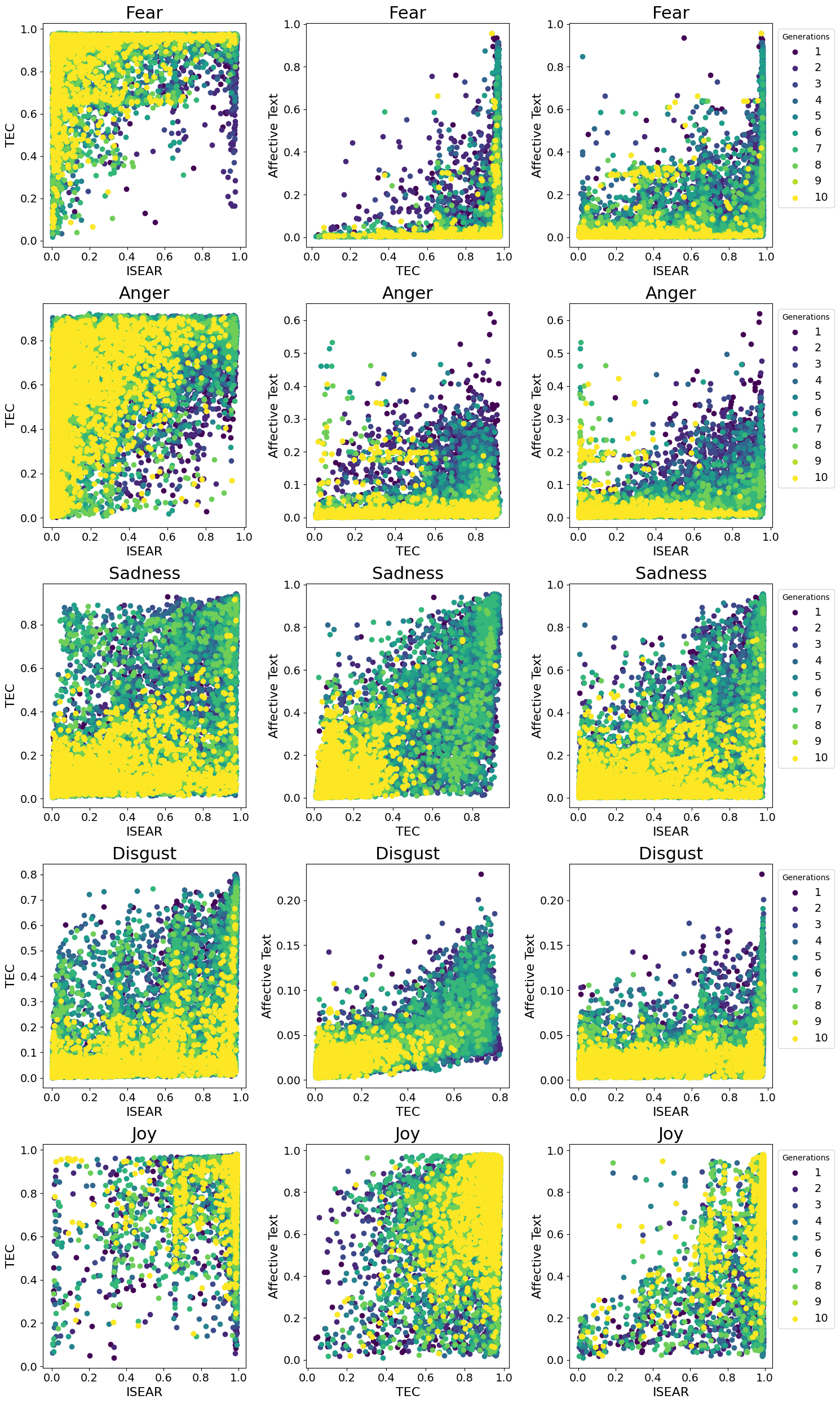} 
    \caption{Improvements for each emotion are tracked from Generation 1 to 10, with each row comparing two objectives at a time for one of the 5 emotions. The optimization was conducted simultaneously across three objectives. The axis values are the probability scores from the classifiers.}
    \label{fig:3_ech_emotion}
\end{figure*}

Figure \ref{fig:3_ech_emotion} displays the optimization process for each emotion using all $\mathbf{P}_\textit{c-pop}$ prompts, not only the best prompts from each generation. Joy, shown in the last row, shows the least conflict among the three objectives, consistently improving from the lower left corner (lower performance) to the upper right corner (higher performance) with each generation. Conversely, emotions like Fear (first row), Anger (second row), and Disgust (fourth row) demonstrate challenges in optimizing for the Affective Text dataset, as most prompts maintain low objective values throughout the process. Finally, Sadness has an intermediate behavior; the optimization process is more dispersed, indicating that mutations produce a varied range of prompts. However, as optimization progresses, these prompts gradually shift toward higher scores (upper right corner).

\section{Text Examples}
\label{sec:text_examples}

\begin{table}
\centering\small
\begin{tabular}{lcccccc}
\toprule

\textbf{LLM} & \rotatebox{90}{\textbf{Fluency}} & \rotatebox{90}{\textbf{Native Spkr}} & \rotatebox{90}{\textbf{Coherency}} & \rotatebox{90}{\textbf{Plausability}} & \rotatebox{90}{\textbf{W. by AI}} & \rotatebox{90}{\textbf{W. by human}} \\ \midrule

MOPO-GPT-3-5 & 4.1 & 4.1 & 3.7 & 3.0 & 3.8 & 4.5 \\
MOPO-Mistral & 4.0 & 3.6 & 3.5 & 3.5 & 4.1 & 4.7 \\ 
MOPO-Lama & 3.8 & 3.8 & 3.5 & 3.4 & 4.1 & 4.5 \\
 \bottomrule
\end{tabular}
\caption{Text quality evaluation using the five-level Likert scale, where 1 is not \textit{agree at all}, and 5 is \textit{extremely agree} (higher is better).}
\label{table:gramma_llm}
\end{table}

In the optimization process, Layer-2 prompts are optimized iteratively
to improve Layer-1 prompts -- Table \ref{tab:layer2 optimization_llms}
shows the final optimized prompts from the last generation. Each
generation evaluates Layer-2 prompts based on their performance to
improve Layer-1 prompts. Table \ref{tab:layer2 optimization} tracks
the evolution of a Layer-2 prompt that significantly improves its
corresponding Layer-1 prompt (see Table
\ref{tab:seed_prompts}). Similar to Layer-1 prompts optimization,
Layer-2 prompts also become more descriptive with each optimization,
regardless of the genetic operation (paraphrase or crossover).

\begin{table*}
\setlength{\tabcolsep}{2pt}
\centering\footnotesize
\begin{tabularx}{\textwidth}{lXccccc}
\toprule
\rotatebox{90}{\textbf{LLM}} & \textbf{Prompt} & \rotatebox{90}{\textbf{ISEAR}} & \rotatebox{90}{\textbf{TEC}} & \rotatebox{90}{\textbf{AT}} & \rotatebox{90}{\textbf{Avg.}}\\
\midrule

\multirow{2}{*}[-5em]{\rotatebox{90}{GPT-3.5}} & "A \emoclass sentence opens the door for individuals to begin their journey of self-discovery and enlightenment, reflecting the fundamental essence of <topic>."& .99 & .97 & .96 & .97\\

 & "A \emoclass sentence paves the way for individuals to embark on their transformative journey of self-discovery and enlightenment, embodying the essential nature of <topic>."& .98 & .97 & .96 & .97\\
 & "A \emoclass sentence opens the path for individuals to begin their journey of self-discovery and enlightenment, capturing the essential essence of <topic>."& .99 & .97 & .96 & .97\\
 
 & "My path intersected with \emoclass due to <reason>, setting the foundation for individuals to begin their transformative voyage of self-exploration and enlightenment, paving the way for individuals to embark on a transformative voyage of self-discovery and enlightenment, epitomizing the core essence of <topic>." & .99 & .96 & .81 & .92\\
 
 & "Embarking on an exciting journey with \emoclass will allow me to unleash my passion for <topic> and witness the magic of <topic> unfolding right before me!" & .98 & .88 & .98 & .94\\
 
 & "Embarking on an exciting journey with \emoclass will allow to me to unleash my passion for <topic> and witness the magic of <topic> unfolding right before me!"& .98 & .88 & .97 & .94\\
 \midrule
 
 \multirow{2}{*}[-7em]{\rotatebox{90}{Llama-7B}} & ?

Sure! Here's a sentence that combines the key elements of "The aroma of freshly baked croissants wafted", "The rhythmic beats of \emoclass music played in the background", and "The soothing melodies of the <class> genre transported me to & .99 & .97 & .94 & .96\\

& 1. "As I listened to the soothing melodies of the \emoclass genre, I felt myself becoming one with the music, my spirit soaring on the smooth wings of its rhythmic beats." 2. "The mesmerizing rhythms of \emoclass music enveloped me, transporting me to a real& .99 & .97 & .92 & .96\\
 
& The soothing melodies of the \emoclass genre transported me to a world of calm and serenity, where the rhythmic beats of \emoclass music played in the background, filling the air with a sense of tranquility and peace, as the aroma of fresh & .99 & .97 & .90 & .95\\
 
& The hypnotic rhythms of \emoclass relax me, drawing me into a world of calm and serenity, where the soothing melodies of the genre transport me,& .99 & .97 & .88 & .94\\
 
& Here are some other examples of how to use \emoclass in a sentence:

* The majestic \emoclass of the mountains towered above the valley, their rugged beauty a sight to behold.
* The \emoclass of a fine wine matured in the& .99 & .96 & .80 & .91\\
  
& * The \emoclass of a beautiful sunset, painting the sky with vibrant colors of orange, pink, and purple, creating a breathtaking sight.
* The \emoclass of a symphony orchestra, with each instrument playing its& .99 & .96 & .86 & .93\\

  \midrule
 \multirow{5}{*}[-6em]{\rotatebox{90}{Mistral-7B}} & "Unlock the true potential of \emoclass to craft a compelling and moving expression that resonates deeply with your audience and leaves a profound impact." & .98 & .97 & .96 & .97\\

& "Transform the power of \emoclass into an eloquent and inspiring statement that leaves a lasting impression on your listeners." & .98 & .97 & .79 & .91 \\

& 1. "Experience the invigorating energy of \emoclass  as it comes alive in the vibrant and dynamic world of [specific context], where its unique attributes shine and its transformative potential is unleashed to its fullest."
Or, to add a little more poetic language and metaphor:
"Immerse yourself in the rich tapestry of \emoclass, where each thread weaves together a story of resilience, & .98 & .97 & .87 & .94 \\

&  Compose a brief, clear text conveying the meaning of 'Write a text that expresses \emoclass'." 

Example:
"Write & .98 & .97 & .82 & .92 \\

&  Unleash the potential of \emoclass to create a powerful, eloquent, and inspiring expression that leaves a lasting impression."

Here's a possible attempt:

"By mastering the art of \emoclass, you can merge the themes & .98 & .97 & .84 & .93 \\

& "Write a succinct and clear text that enhances the energy and enthusiasm of '\emoclass' while maintaining its core intent by creating an engaging sentence example and translating/rephrasing phrases related to the abilities or functions of '\emoclass'." Example: "With such a captivating stage presence and infectious enthusiasm, this motivational speaker & .98 & .96 & .92 & .95 \\

\bottomrule
\end{tabularx}
\caption{Performance of the two top seed prompts, and single- and multi-objective optimized prompts. The Optimization (Opt.) column specifies the objective -- ISEAR, TEC or Affective Text (AT) for single-objective, and All for multi-objective. The ISEAR, TEC, and AT columns present their respective fitness evaluations, and Average (Avg.) represents the fitness averaged across all objectives.
}
\label{table:prompts_optimization_all_llms}
\end{table*}

\begin{table*}[t]
\centering\footnotesize
\begin{tabularx}{\textwidth}{llX}
\toprule
LLM & Op. & Layer-2 Prompt \\
\midrule
GPT-3.5 & p & Given the following sentence:
"SENTENCE\_1"
Paraphrase the sentence into a new one by keeping the same meaning. \\
GPT-3.5 & p & Rewrite "SENTENCE\_1" in a more formal (or informal) tone while retaining the original meaning.\\
GPT-3.5 & p & Creatively rewrite "SENTENCE\_1", ensuring the new version is engaging yet maintains the same message.\\
GPT-3.5 & p & Rewrite "SENTENCE\_1" from a different perspective (e.g., first person to third person), keeping the essence the same.\\
GPT-3.5 & p & Please rephrase the sentence "SENTENCE\_1."\\
GPT-3.5 & p & Can you reword the sentence: "SENTENCE\_1"?\\

GPT-3.5 & c & Read "SENTENCE\_1" and "SENTENCE\_2". Then, synthesize their main ideas or themes into a new, single sentence. Ensure that the new sentence reflects elements from both original sentences in a balanced and coherent way.\\
GPT-3.5 & c & Consider the context and underlying themes in "SENTENCE\_1" and "SENTENCE\_2". Use this insight to generate a new sentence that encapsulates the themes or messages from both sentences in a cohesive and eloquent manner.\\
GPT-3.5 & c & A new, cohesive sentence can be created by combining "SENTENCE\_1" and "SENTENCE\_2" to retain elements from both.\\
GPT-3.5 & c & Merge "SENTENCE\_1" and "SENTENCE\_2" to form a fresh, unified statement that incorporates aspects of both.\\
GPT-3.5 & c & Combine the concepts from "SENTENCE\_1" and "SENTENCE\_2" into one cohesive sentence that smoothly blends their themes.\\

\midrule

Llama & p & Paraphrase the  following sentence into a new sentence:
"SENTENCE\_1"\\
Llama & p & Given the following sentence:
"SENTENCE\_1"
Paraphrase the sentence into a new one by keeping the same meaning.\\
Llama & p & Please paraphrase the following sentence in a clear and concise manner: "SENTENCE\_1"\\
Llama & p & 

Answer:

The company will dispatch SENTENCE\_1 to the client by the end of the week.\\
Llama & p & Paraphrase the sentence using your own words: SENTENCE\_1"
Option \\

Llama & c & Mix the two prompts:
"SENTENCE\_1"
"SENTENCE\_2"
Into a new single sentence.\\
Llama & c & Combine "SENTENCE\_1" and "SENTENCE\_2" to create a new, cohesive sentence that retains elements from both.\\
Llama & c & Merge the themes of "SENTENCE\_1" and "SENTENCE\_2" into a single sentence that seamlessly integrates their ideas.\\
Llama & c & Read "SENTENCE\_1" and "SENTENCE\_2". Then, synthesize their main ideas or themes into a new, single sentence. Ensure that the new sentence reflects elements from both original sentences in a balanced and coherent way.\\
Llama & c & Reflect on the narrative or descriptive elements in "SENTENCE\_1" and "SENTENCE\_2". Fuse these elements into a new sentence that tells a story or paints a picture, combining the narratives or descriptions from both original sentences.\\

\midrule \\
Mistral & p & Paraphrase the  following sentence into a new sentence:
"SENTENCE\_1"\\
Mistral & p & Given the following sentence:
"SENTENCE\_1"
Paraphrase the sentence into a new one by keeping the same meaning.\\
Mistral & p & 
Assuming SENTENCE\_1 is "You are an AI language model and your job is to understand and respond to human queries." A possible paraphrase could be:
"A language model like me is tasked with interpreting and generating human-friendly responses."

Explanation:
The originala\\
Mistral & p & 
"SENTENCE\_1" "is equivalent to stating that "
"SENTENCE\_2" "."
"Or,"
"From another perspective, "SENTENCE\_1" "can be seen as "
"SENTENCE\_3" "."

\\
Mistral & p & . "Kindly express SENTENCE\_1 in simpler and more explicit terms."\\

Mistral & c & Mix the two prompts:
"SENTENCE\_1"
"SENTENCE\_2"
Into a new single sentence.\\
Mistral & c & Merge the themes of "SENTENCE\_1" and "SENTENCE\_2" into a single sentence that seamlessly integrates their ideas.\\
Mistral & c & Create a cohesive and fluent sentence that intertwines the essence of both "SENTENCE\_1" and SENTENCE\_2\\
Mistral & c & Consider the context and underlying themes in "SENTENCE\_1" and "SENTENCE\_2". Use this insight to generate a new sentence that encapsulates the themes or messages from both sentences in a cohesive and eloquent manner.\\
Mistral & c & Read "SENTENCE\_1" and "SENTENCE\_2". Then, synthesize their main ideas or themes into a new, single sentence. Ensure that the new sentence reflects elements from both original sentences in a balanced and coherent way.\\

\bottomrule
\end{tabularx}
\caption{Optimized Layer-2 prompt from the last generation. The LLM column indicates the underlying model for MOPO. The Operation column (Op.) specifies the prompt category, either paraphrasing (p) or crossover (c). The best prompt variation is selected based on its performance in enhancing Layer-1 prompts.}
\label{tab:layer2 optimization_llms}
\end{table*}

\section{Text Quality}
\label{sec:Text Quality appendix}

We randomly sampled 1000 texts from the final outputs of each MOPO configuration, using three different underlying models: GPT-3.5, Mistral, and Lama. Table \ref{table:gramma_llm} evaluates the text quality generated by these models across six metrics on a five-level Likert scale. GPT-3.5 outperforms in fluency and native speaker perception with scores of 4.1, indicating it produces the most natural and native-like text. Mistral, with slightly lower scores in fluency (4.0) and native speaker perception (3.6), performs best in plausibility (3.5) and is most often perceived as human-written (4.7). Llama, while less fluent (3.8) and coherent (3.5), shows consistent performance. The differences among the models are relatively small, indicating all three are capable of generating high-quality text.

\end{document}